\documentclass[10pt,twocolumn,letterpaper]{article}

\usepackage[pagenumbers]{cvpr} 

\usepackage{graphicx}
\usepackage{amsmath}
\usepackage{amssymb}
\usepackage{booktabs}
\usepackage[table,xcdraw]{xcolor}
\usepackage{enumitem}
\usepackage{caption}
\usepackage{multirow}
\usepackage[accsupp]{axessibility}
\newcommand{\mcircle}[1]{\raisebox{1pt}{\textcircled{\raisebox{-.9pt} {#1}}}}
\newlength\savewidth

\definecolor{cvprblue}{rgb}{0.21,0.49,0.74}
\definecolor{lightblue}{rgb}{0.68, 0.85, 0.9}

\usepackage[pagebackref,breaklinks,colorlinks,citecolor=cvprblue]{hyperref}

\usepackage[capitalize]{cleveref}
\crefname{section}{Sec.}{Secs.}
\Crefname{section}{Section}{Sections}
\Crefname{table}{Table}{Tables}
\crefname{table}{Tab.}{Tabs.}

\def\cvprPaperID{12758} 
\def\confName{CVPR}
\def\confYear{2024}

\begin{document}

\title{Improved Zero-Shot Classification by Adapting VLMs with Text Descriptions}


\author{Oindrila Saha \quad \quad Grant Van Horn \quad \quad  Subhransu Maji\\
University of Massachusetts, Amherst\\
{\tt\small \{osaha, gvanhorn, smaji\}@umass.edu}}


\maketitle
\begin{abstract}
The zero-shot performance of existing vision-language models (VLMs) such as CLIP~\cite{radford2021learning} is limited by the availability of large-scale, aligned image and text datasets in specific domains. In this work, we leverage two complementary sources of information---descriptions of categories generated by large language models (LLMs) and abundant, fine-grained image classification datasets---to improve the zero-shot classification performance of VLMs across fine-grained domains. On the technical side, we develop methods to train VLMs with this ``bag-level" image-text supervision. We find that simply using these attributes at test-time does not improve performance, but our training strategy, for example, on the iNaturalist~\cite{van2018inaturalist} dataset, leads to an average improvement of 4-5\% in zero-shot classification accuracy for novel categories of birds~\cite{wah2011caltech} and flowers~\cite{nilsback2008automated}. Similar improvements are observed in domains where a subset of the categories was used to fine-tune the model. By prompting LLMs in various ways, we generate descriptions that capture visual appearance, habitat, and geographic regions and pair them with existing attributes such as the taxonomic structure of the categories. We systematically evaluate their ability to improve zero-shot categorization in natural domains. Our findings suggest that geographic priors can be just as effective and are complementary to visual appearance. Our method also outperforms prior work on prompt-based tuning of VLMs. We release the benchmark, consisting of 14 datasets at \url{https://github.com/cvl-umass/AdaptCLIPZS}, which will contribute to future research in zero-shot recognition.
\end{abstract}
\section{Introduction}
\label{sec:intro}
Recent improvements in zero-shot classification have been due, in part, to success in training VLMs at scale. Models such as CLIP~\cite{radford2021learning}, ALIGN~\cite{jia2021scaling}, and BLIP~\cite{li2022blip} use massive datasets of image and text pairs to learn a common embedding between visual and natural language domains. However, we find existing VLMs show poor performance in encoding visual attributes in fine-grained domains, beyond simply recognizing the name of the category. For example, we observe concatenating visual attribute descriptions to the species name for the bird species in the
CUB~\cite{wah2011caltech} dataset improves the zero-shot classification from 50.5\% to only 50.7\%, while for Cars~\cite{krause20133d}, the performance even drops slightly (see Tab.~\ref{tab:maintable}). Although the datasets on which VLMs are trained are extensive, they often lack the details that experts may require for fine-grained categorization. At the same time, collecting large-scale image-caption datasets in these domains requires significant effort, making training similar models challenging.

\begin{figure}
    \centering
    \includegraphics[width=\linewidth]{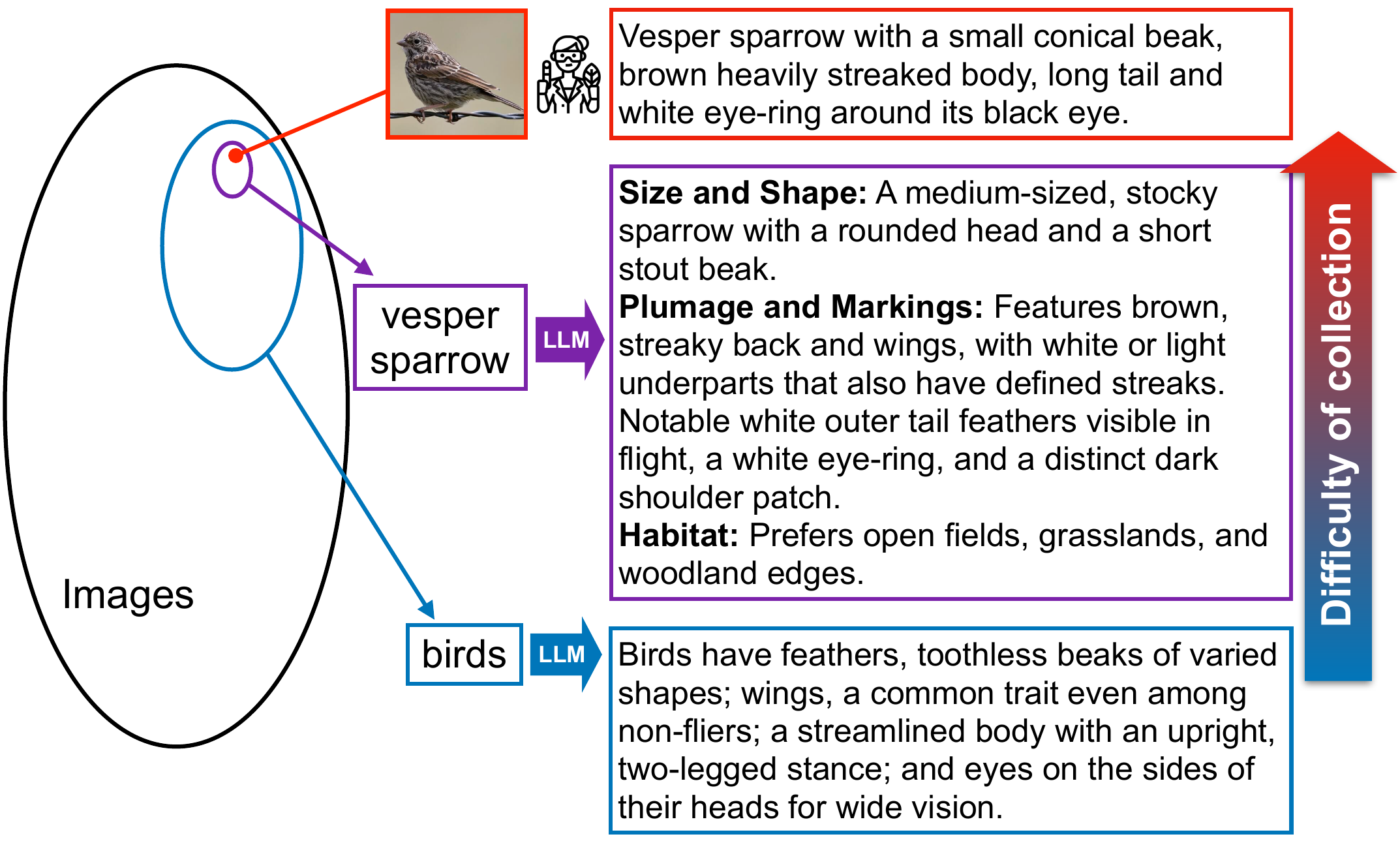}
    \caption{\textbf{Motivation.} Collecting image captions in fine-grained domains requires expertise (top row), but LLMs can generate structured (e.g., shape or appearance) and accurate descriptions of categories at both the coarse (e.g., birds) and fine-grained level (e.g., Vesper Sparrow). Rich descriptions of fine-grained categories can be paired with existing datasets, such as iNaturalist~\cite{van2018inaturalist} and NABirds~\cite{van2015building} to generate coarsely-aligned image-text datasets for fine-tuning VLMs. This improves their zero-shot performance on a range of benchmarks, generalizing to novel categories and tasks.}
    \label{fig:splash}
    \vspace{-2mm}
\end{figure}

In this work, we leverage two complementary sources of information—large language models (LLMs) and abundant, fine-grained image classification datasets—to improve the zero-shot classification performance of VLMs across a variety of fine-grained domains. Concretely, we generate large datasets of images aligned with text by pairing images within a category with descriptions of that category generated by LLMs as seen in Fig.~\ref{fig:splash}. We find this approach works well, as images within a fine-grained domain share many attributes, unlike in coarse categories with larger intra-category variation. At the same time, we find that LLMs are capable of accurately describing appearance, habitat, and other properties for a wide range of categories, allowing us to systematically generate datasets in a scalable manner. In other words, fine-grained labels allow us to bridge the gap between image-level captions required for VLM training, and general information about visual categories contained in LLMs.

\begin{figure}
    \centering    
    \includegraphics[width=\linewidth]{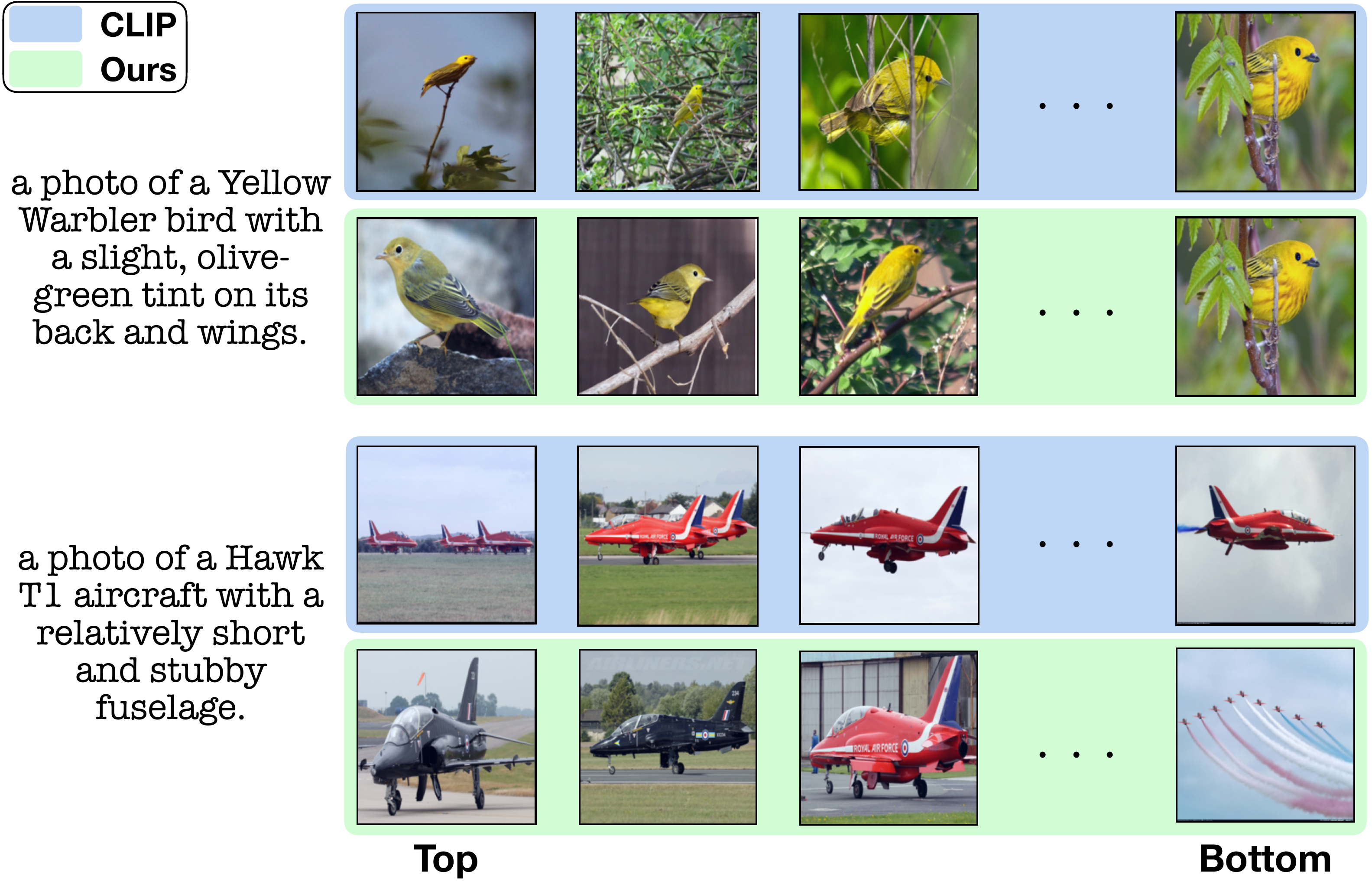}
    \captionsetup{font=footnotesize}
    \caption{\textbf{Visualizing image-text similarity.} All images within a category are sorted in order of similarity to a given text predicted by CLIP and our fine-tuned CLIP\textsuperscript{FT}+A. For example, our method identifies birds which show olive-green tint on their back as the top images, whereas CLIP selects birds with visibly brown upperparts or occluded back. The image with lowest similarity which has the occluded back remains the same for both models, showing our model does not learn incorrect attribute associations even though we stochastically pair every attribute with every image during training. On the aircraft example our model predicts higher similarity to images with prominently visible fuselage. CLIP identifies the least similar image as one in which fuselage is visible, but ours chooses one where aircrafts are too far to make out the shape of fuselage.\vspace{-2mm}}
    \label{fig:viz}
\end{figure}

On the technical side, we develop methods to train VLMs with ``bag-level" supervision.
In our dataset a set of images are grouped with a set of descriptions and lack the image-text correspondences. Some of the descriptions may not apply to an image (e.g., the part may be occluded). However, we find that training by stochastically pairing the images and text within a category, followed by a category-level contrastive loss similar to CLIP objective offers robust improvements in performance. Adapting semi-supervised learning approaches such as FixMatch~\cite{sohn2020fixmatch} or Knowledge distillation~\cite{chen2020big} results in minor improvements (\S~\ref{sec:alttrain}). A detailed investigation of the image-text association within a category suggests the model is able to correctly associate the visual attributes with the corresponding text even they are paired randomly during training (Fig.~\ref{fig:viz}).

We systematically evaluate the effectiveness of our method by assessing the zero-shot classification performance on novel classes. We find that simply using these attributes of novel classes generated by LLMs does not improve performance when using CLIP (Tab.~\ref{tab:maintable}). However, our training strategy leads to an average improvement of 4-5\% in accuracy across 12 datasets, and outperforms baselines (Tab.~\ref{tab:compare}). For natural domains (e.g., iNaturalist and NABirds), we prompt LLMs in various ways to generate descriptions that capture visual appearance, habitat, and geographic regions, and pair them with existing attributes within the dataset, such as taxonomic structure. Our results indicate that geographic priors are equally effective and complementary to visual appearance cues (see Tab.~\ref{tab:othertexts}). Training on iNaturalist without any bird classes improves the performance of CLIP on CUB by more than 3\%, and we observe similar improvements when evaluating across other domains (see Tab.~\ref{tab:crossdomain}). Improvements are consistent across text generated by different LLMs, as well as by humans (see Tab.~\ref{tab:otherllm}). Our model also results in relative error reduction of 4.1\% over CLIP on the challenging NeWT dataset~\cite{van2021benchmarking}.


\section{Related Work}
\label{sec:related work}

\noindent\textbf{Zero-shot Image Classification using VLMs.}
Vision Language models (VLMs)~\cite{radford2021learning, jia2021scaling, singh2022flava, yuan2021florence, furst2022cloob} learn to associate images with their corresponding text captions. Learning a shared embedding makes them perform exceptionally well in zero-shot classification tasks when paired with an appropriate text such as the class name during test time. FLAVA~\cite{singh2022flava} learns using paired as well as unpaired images and texts using different losses for multimodal and unimodal understanding. ALIGN~\cite{jia2021scaling} uses a large number of noisy image-text data by obtaining alt-texts for images and trains using a contrastive loss. CLIP~\cite{radford2021learning} is trained on a smaller and cleaner dataset of image-text pairs using a similar objective function. It employs a vision model and a language model to learn joint embeddings of images and text. While training, it maximizes the similarity between related image-text pairs and minimizes similarity between unrelated pairs. At test time the similarity over captions such ``a photo of a \texttt{[class name]}" over all classes in the domain for each image is found. The image is classified to the class with the caption with highest probability. The original paper shows that manual prompt tuning can boost zero-shot classification accuracy.

\noindent\textbf{Generating Better Prompts.} Prior work on prompt tuning~\cite{khattak2023maple, zhu2023prompt, shu2022test, gan2023decorate, jia2022visual} has focused on improving the text descriptions of classes. For example, CoOp~\cite{zhou2022learning} appends learnable context vectors to the class name texts to improve classification. CoCoOp~\cite{zhou2022conditional} and related methods have also explored prompting the vision encoder simultaneously. While prompt tuning has proven useful for adapting models to a set of ``base categories", its performance on novel categories still falls short of the CLIP baseline.
 
Another line of research aims at querying LLMs to generate prompts or attributes of categories. CHiLS~\cite{novack2023chils} refines classes based on GPT descriptions (e.g., taxonomic structure) and maps the image to one of the subcategories to improve classification. We also explore the ability to learn taxonomy-based attributes in our work. Menon \etal~\cite{menon2022visual} and CuPL~\cite{pratt2023does} append class-specific attributes obtained from GPT~\cite{brown2020language} to simple prompts, e.g. ``a photo of a \texttt{[class]}" to improve zero-shot performance at test-time, similar to our approach. However, we find that CLIP struggles to recognize nuanced attributes in fine-grained domains where we observe little to no improvement in classification performance, motivating the need for fine-tuning VLMs.

\noindent\textbf{Fine-tuning VLMs.} Most work on fine-tuning VLMs has focused on parameter efficient updates using lightweight adapters~\cite{pantazis2022svl, peng2023sgva, zhang2021tip, gao2023clip, zhang2023LLaMA, maniparambil2023enhancing} for improving few-shot classification. For example, CLIP adapter~\cite{gao2023clip} trains a few learnable layers on top of the encoders, while Maniparambil \etal~\cite{maniparambil2023enhancing} query LLMs for class-wise descriptions for tuning an external adapter network. Their work improves over CoCoOp and CLIP adapter for unseen classes, however, as before, most approaches show no improvement over CLIP, especially on novel classes in fine-grained datasets. 

Another line of research~\cite{zhang2021domain, goyal2023finetune, singha2023ad, wortsman2022robust, tian2022vl} involves fine-tuning CLIP for robustness to domain shifts. For instance, WiSE-FT~\cite{wortsman2022robust} utilizes weight-space ensembling to improve performance on a sketch version of ImageNet. Similarly,  LaFTer~\cite{mirza2023lafter} employs fine-tuning both the image and text encoders using unpaired images and texts obtained by querying LLMs. These techniques focus on adapting to a target distribution, such as a specific set of test images or classes, rather than on generalizing to novel classes.

Two works similar to ours, GIST~\cite{lewis2023gist} and I2MVFormer~\cite{naeem2023i2mvformer}, also utilize GPT to generate category-specific texts for fine-grained domains. GIST pairs each image with the $n$ most similar texts within the category based on CLIP similarity and consolidates them into a caption. In contrast, our approach stochastically pairs images with text, a strategy we found to be more robust. Notably, our experiments showed that biasing sampling towards similar image and text pairs led to inferior results (see \S~\ref{sec:alttrain}). I2MVFormer uses a LLM to generate class descriptions based on texts provided by annotators and Wikipedia documents. The key differences between their work and ours lie in the use of human effort in their training data generation and starting from scratch resulting in much lower performance compared to ours. Finally, in contrast to both these works, we explore training using visual, habitat, and location information, as well as training on larger datasets such as NABirds and iNaturalist.




\paragraph{Summary} To the best of our knowledge, ours is the first method demonstrating that fine-tuning CLIP with class-specific descriptions obtained by querying LLMs improves the zero-shot performance in fine-grained domains. Our approach leverages LLMs to generate image-text data that are coarsely aligned, making it particularly effective for fine-grained categories.  Moreover, unlike prior work~\cite{menon2022visual,maniparambil2023enhancing}, our method queries LLMs along various dimensions such as visual, taxonomy, habitat and geographic priors, and systematically evaluates their effectiveness. 


\begin{figure*}
    \centering
    \includegraphics[width=1.0\linewidth]{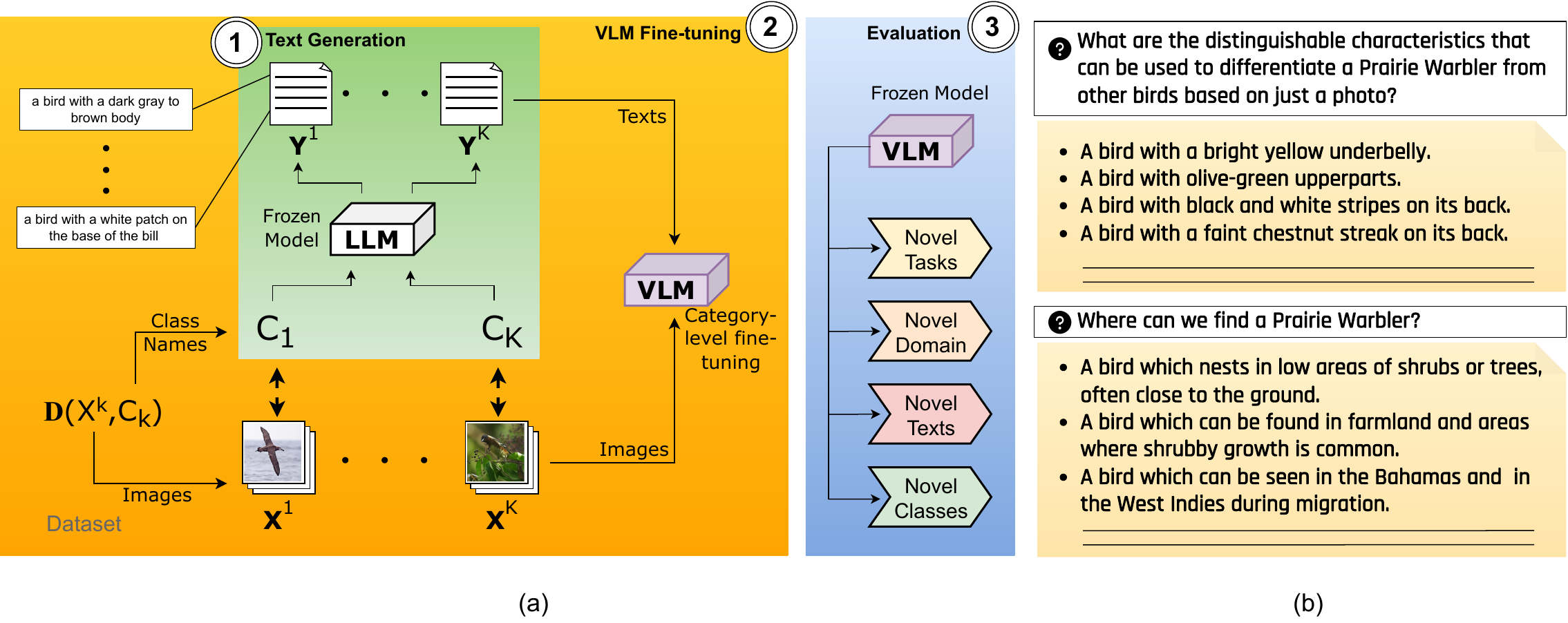}
    \vspace{-7mm}
    \caption{\textbf{Fine-tuning VLMs to improve zero-shot performance.} a) Our framework for \mcircle{1} generating fine-grained attributes per class using LLMs, \mcircle{2} category-level fine-tuning of VLMs and \mcircle{3} evaluating on a series of challenging unseen scenarios. b) We show examples of texts produced in step \mcircle{1}.\vspace{-5mm}}
    \label{fig:method}
\end{figure*}

\section{Method}
Consider a dataset $\mathcal{D}=\{(x_i,y_i)\}_{i=1}^n$ consisting of images $x_i \in {\cal X}$ and labels $y_i \in {\cal Y}$. A VLM such as CLIP~\cite{radford2021learning} consists of an image encoder $\Theta$ and a text encoder $\Phi$ such that $\Theta(x) \approx \Phi(y)$ for images $x$ with label $y$. We want to improve the \textbf{zero-shot performance} of CLIP on novel categories in fine-grained domains by fine-tuning the image and the text encoders. We do so either by splitting a dataset $\mathcal{D}$ into $\mathbf{K_{train}}$ training and $\mathbf{K_{test}}$ testing classes, or train our model on large datasets such as iNaturalist and NABirds by excluding classes or domains overlapping with our test set. Our framework consists of: 1) generating textual descriptions given the class names by prompting LLMs in different ways (\S~\ref{sec:queryllm}); 2) fine-tuning CLIP using these descriptions using our proposed approach (\S~\ref{sec:finetune}); and 3) evaluating the models on downstream tasks (\S~\ref{sec:eval}). Fig.~\ref{fig:method} provides an illustration of our method.





\subsection{Dataset Generation}
\label{sec:queryllm}
For each dataset we generate texts for every class which can be used to differentiate it from other classes in the domain using \textbf{visual attributes}. We query an LLM as:

\begin{quote}
    \makebox[\linewidth]{%
        \colorbox{lightblue}{%
            \hspace*{0mm} 
            \begin{minipage}{\dimexpr\linewidth+10\fboxsep\relax} 
                \fontsize{9pt}{10pt}\selectfont 
                What characteristics can be used to differentiate \texttt{[class]} from other \texttt{[domain]} based on just a photo? Provide an exhaustive list of all attributes that can be used to identify the \texttt{[domain]} uniquely. Texts should be of the form ``\texttt{[domain]} with \texttt{[characteristic]}".
            \end{minipage}%
        }%
    }
\end{quote}
Here \texttt{[class]} is the class name for the $K$ classes in the dataset \texttt{[domain]}. Each domain is associated to different datasets, for example for the CUB200 dataset, \texttt{[domain]} is ``bird" and for iNaturalist dataset it is ``organism". Appending the domain \texttt{[domain]} is helpful because it provides context about the set of the other classes to distinguish from, and reduces confusion across similar class names in other domains.  The LLM produces $l_k$ descriptions for category $k$. We append ``a photo of \texttt{[class]}" to the generated texts resulting in descriptions of the form `` a photo of a \texttt{[class]} \texttt{[domain]} with \texttt{[characteristic]}" for each class (more details in the experiment section). This results in a set of descriptions $\mathbf{Y^{k}}$ for each category.

We also separately query about the \textbf{habitat and geographic location} of occurrence for the classes in CUB200, Flowers102, NABirds and iNaturalist datasets. For this purpose we use the prompt:


\begin{quote}
    \makebox[\linewidth]{%
        \colorbox{lightblue}{%
            \hspace*{0mm} 
            \begin{minipage}{\dimexpr\linewidth+10\fboxsep\relax} 
                \fontsize{9pt}{10pt}\selectfont 
                Where can we find a \texttt{[class]}? Produce a list of habitat and geographic location information that can be used to identify the \texttt{[domain]}.
            \end{minipage}%
        }%
    }
\end{quote}
We add the texts obtained to the category-level corpus $\mathbf{Y^k}$. The impact of these location-specific texts and the improvements offered are described in the results section. Examples of visual and habitat descriptions are in the Appendix. We determine the \textbf{correctness of the texts produced} for 4-6 classes of CUB, Aircraft and Cars by manual fact-checking texts with the help of online sources. We find that 96$\%$ (CUB), 90$\%$ (Aircraft) and 96$\%$ (Cars) are marked as correct by study participants (more details in Appendix). 
However, a challenge is that this manual vetting does not scale to large datasets, and we therefore rely on empirical results to support the utility of the generated text.

\subsection{VLM Fine-Tuning}
\label{sec:finetune}
CLIP~\cite{radford2021learning} is trained with image caption pairs.  However, in our case we have a set of images $\mathbf{X^{k}}$ and a set of texts $\mathbf{Y^{k}}$ for classes $k \in \mathbf{K_{train}}$ in our training set. We address this by pairing every image with randomly sampled text from the corresponding category during training. However, we cannot directly use the batch-level cross-entropy loss used by CLIP which treats the paired text as positive and rest of the texts as negative. This is because the same batch can contain multiple pairs with images and texts belonging to the same category.
Below we describe our modification of the objective function that addresses this.

In each iteration of training we sample a batch of size $N$ consisting of $\{(x_i,y_i)\}_{i=1}^N$ pairs where both $x_i$ and $y_i$ is an image and text from the same class. Let the similarity score obtained using the forward pass of CLIP for image $x_i$ and text $y_j$ be $S_{ij}$. Let $\text{c}(i)$ be the category of image-text pair $(x_i,y_i)$.
Let $G_{i} = \{ j \mid \text{c}(j) = \text{c}(i)\}$ denote the indices of pairs that belong to the same class as pair $(x_i,y_i)$.
Then the loss function for images is:
\begin{align}
    \mathcal{L}_{image} &= - \frac{1}{N} \sum_{i=1}^{N} \frac{1}{|G_i|} \sum_{j \in G_i} \log \frac{\exp(S_{i,j}/\tau)}{\sum_{r=1}^{N} \exp(S_{i,r}/\tau)}
\end{align}
and the corresponding one for texts is:
\begin{align}
    \mathcal{L}_{text} &= - \frac{1}{N} \sum_{j=1}^{N} \frac{1}{|G_j|} \sum_{i \in G_j} \log \frac{\exp(S_{i,j}/\tau)}{\sum_{r=1}^{N} \exp(S_{r,j}/\tau)}
\end{align}
where $\tau$ is a learnable temperature parameter. The overall loss for fine-tuning is:
\begin{align*}
    \mathcal{L}_{ft} = \mathcal{L}_{image} + \mathcal{L}_{text}
\end{align*}
The objective aggregates the image text similarity across all image and text pairs from the same category within the batch. To avoid overfitting on small datasets we maintain momentum encoders whose weights ($\theta_{EMA}$, $\phi_{EMA}$) are updated with the exponential moving average (EMA) of the weights of the encoders ($\theta$, $\phi$) which is trained using the objective $\mathcal{L}_{ft}$:
\begin{align*}
    \theta_{EMA} &\leftarrow m\theta_{EMA} + (1-m)\theta_{E}\\
    \phi_{EMA} &\leftarrow m\phi_{EMA} + (1-m)\phi_{E},
\end{align*}
where $m$ is a momentum parameter. 
All encoders are initialized using the pre-trained weights of CLIP. 

\subsection{Evaluation for Zero-shot Classification}
\label{sec:eval}
To evaluate a model on unseen classes we similarly query the LLM as described in \S\ref{sec:queryllm} to obtain texts $\mathbf{Y^{k}}$, for $k \in \mathbf{K_{test}}$. For any given image $x$ we can find the similarity score using a VLM for every text $y^k_m $ for $m \in \{1, \dots, l_k\}$ and $k \in \mathbf{K_{test}}$. Denote the similarity between image $x$ and text $y^k_m$ as $S^k_m$. The predicted class is:

\begin{align}
\operatorname*{argmax}_{k} \frac{1}{l_k} \sum_{m=1}^{l_k} \frac{\exp(S^k_m)}{\sum_{p \in \mathbf{K_{test}}}\sum_{q=1}^{l_p} \exp(S^p_q)}
\end{align}
The score represents average similarity between an image and the texts corresponding to each class. Our initial experiments suggested that simple averaging of probabilities is more robust than alternatives such as the geometric mean.

\begin{table*}[]
\centering
\begin{tabular}{l|lllll}
Methods & CUB   & Stanford Cars & FGVC Aircrafts & Flowers 102 & Food 101  \\ \hline
CLIP  & 50.54 & 69.72         &   29.27        & 71.78   & 88.32   \\
CLIP + A  & 50.71 & 69.47         &   30.35        & 75.37   & 90.08   \\
CLIP\textsuperscript{FT} & 50.81 $\pm$ 0.04 & 69.61 $\pm$ 0.07          &    31.10 $\pm$ 0.02            &    73.68 $\pm$ 0.00    &    88.32 $\pm$ 0.00      \\
\rowcolor[HTML]{B6D7A8} 
CLIP\textsuperscript{FT} + A & \textbf{53.34} $\pm$ 0.08  & \textbf{71.63} $\pm$ 0.06  &\textbf{36.41} $\pm$ 0.02 & \textbf{77.05} $\pm$ 0.00 &  \textbf{93.71} $\pm$ 0.01              
\end{tabular}
\caption{\textbf{Comparison with CLIP ViT-B/32 on zero-shot performance on fine-grained domains.} We compare our method CLIP\textsuperscript{FT} + A to the baselines defined in \S~\ref{sec:baselines}. We significantly improve over baseline CLIP evaluated with both ``a photo of a \texttt{[class]} \texttt{[domain]}" and LLM attributes. We also fine-tune CLIP with only ``a photo of a \texttt{[class]} \texttt{[domain]}" text and compare with our method to show that our improvements are not due to seeing domain-specific images but also by learning correlations between images and fine-grained attributes.\vspace{-3mm}}
\label{tab:maintable}
\end{table*}

\begin{table*}[]
\centering
\begin{tabular}{l|cccccc}
 & Stanford Cars & FGVC Aircrafts & Flowers 102 & EuroSAT & Food 101 &  ImageNet \\ \hline
CLIP & 74.94 & 36.47 & 77.05 & 64.05 & 92.49 &  67.41  \\
CLIP + A & 73.83 & 36.47 & 80.84 & 71.51 & 93.72 & 69.74 \\
CLIP-A-self~\cite{maniparambil2023enhancing} & 72.90 & 33.00 & 75.30 & 70.50 & 91.20 &  68.30  \\
\rowcolor[HTML]{B6D7A8} 
CLIP\textsuperscript{FT} + A & \textbf{75.78} & \textbf{40.75} & \textbf{81.26} & \textbf{81.82} & \textbf{95.08} & \textbf{71.87}  
\end{tabular}
\caption{\textbf{Comparison to prior work using ViT B/16 architecture on zero-shot classification.} We show that across a variety of datasets from finer to coarser domains we considerably boost performance over baselines. Here we train using only 16 images per class and test on \textbf{unseen classes} for fair comparison to CLIP-A-self. We do not compare on CUB dataset as CLIP-A-self uses a 3:1 split on CUB, whereas we use 1:1 across all datasets. \vspace{-5mm}}
\label{tab:compare}
\end{table*}

\begin{table}[]
\centering

\begin{tabular}{lcc}
Texts                        & CUB            & Flowers 102 \\ \hline
Visual                       & 53.34          & 77.05     \\
Taxonomy                     & 53.07          &   -       \\
Habitat                     & 53.69          & 76.00         \\
Vis. + Hab.             & 54.01          &      77.22     \\
Vis. + Tax. + Hab. & \textbf{54.23} &   -       
\end{tabular}
\caption{\textbf{Evaluating CLIP\textsuperscript{FT} + A using different types of text.} We query LLMs to produce visual (vis.) and habitat (hab.) information separately and use taxonomy (tax.) information available with dataset. We train with the type of text specified in each row and test with the same type. Using habitat information works slightly better than using visual information for CUB. All three types of texts are complementary.\vspace{-3mm}}
\label{tab:othertexts}
\end{table}

\begin{table}[]
\centering
\centering
\begin{tabular}{lcc}
Train Set                        & CUB & Flowers 102 \\ \hline
NABirds\textbackslash overlap &  55.32   & -       \\
iNat\textbackslash overlap    & 54.58 &  77.05   \\
iNat\textbackslash Birds       &  53.89   & -       \\
iNat\textbackslash Plants     & -   &    76.63    
\end{tabular}
\caption{\textbf{Evaluating domain transfer performance.} Our method offers substantial gain over baseline CLIP (Tab.~\ref{tab:maintable}) even when trained on external datasets. Performance boost is competitive even when removing all bird or plant classes from iNat to test on CUB and Flowers respectively.\vspace{-3mm}}
\label{tab:crossdomain}
\end{table}

\begin{table}[]
\begin{tabular}{l|cc}
Testing Texts        & CLIP + A & CLIP\textsuperscript{FT} + A       \\ \hline
\texttt{[class]} \texttt{[domain]} & 50.54            & 52.29 \\\hline
GPT 4 Vis.    & 50.71            & 53.34 \\
GPT 3 Vis.    &  51.08        

& 53.35  \\
LLaMA Vis.    &  50.10               &    52.52            \\
Ground Truth Vis. & 52.53 & 53.99 \\\hline
GPT 4 Vis. + Tax. + Hab.            & 52.83            & 54.23 \\

GPT 3 Vis. + Tax. + Hab.            &  52.63                &  53.58           \\

LLaMA Vis. + Tax. + Hab.            &   50.85              &   52.64    \\        

\end{tabular}
\caption{\textbf{Evaluating model trained using GPT4 with texts obtained from other models.} Our model consistently improves over pre-trained CLIP when evaluated with texts obtained from different LLM models (GPT3.5 and LLaMA2-7B) as well as GT aggregated captions and ``a photo of a \texttt{[class]} \texttt{[domain]}".\vspace{-3mm}}

\label{tab:otherllm}
\end{table}

\section{Experiments}
In this section we present the experimental details of our approach. We outline the datasets we use, the particulars of implementation for each part of the method as well as the details of the baselines we compare our method to.

\subsection{Datasets}
We use a variety of fine-grained classification datasets including \textbf{CUB~\cite{wah2011caltech}} (200 classes), \textbf{Flowers 102~\cite{nilsback2008automated}} (102 classes), \textbf{Stanford Cars~\cite{krause20133d}} (196 classes), \textbf{FGVC Aircrafts~\cite{maji13fine-grained}} (100 classes) and \textbf{Food101~\cite{bossard14}} (101 classes). We also apply our method on some coarser datasets including \textbf{EuroSAT~\cite{helber2017eurosat}} (10 classes), \textbf{ImageNet~\cite{ILSVRC15}} (1000 classes), \textbf{CalTech101~\cite{FeiFei2004LearningGV}} (100 classes), \textbf{DTD~\cite{cimpoi14describing}} (47 classes), \textbf{Oxford Pets~\cite{parkhi2012cats}} (37 classes), \textbf{Sun397~\cite{Xiao:2010}} (397 classes) and \textbf{UCF101~\cite{soomro2012ucf101}} (101 classes). For all these datasets, we use the first half of the classes (ordered by ids of the original dataset) for training and second half for zero-shot testing. 

We also use \textbf{NABirds~\cite{van2015building}} which contains 404 bird classes at species level. We remove the overlapping classes of the CUB testing set from these to obtain 331 training classes. 
Along with train and test classes being different, this setting also represents a \textit{distribution shift} in the images of training and testing as images for CUB and NABirds have been obtained in different manners.

\textbf{iNaturalist~\cite{van2018inaturalist}} 2021 is another dataset we utilize to illustrate that our method scales and generalizes. iNaturalist contains 10k classes belonging to 11 general categories (such as birds, plants, fishes). First, in a similar setting to NABirds we remove overlapping test classes of CUB to train a model for testing on CUB. Secondly, we remove all bird classes from iNat and train a model on the remaining classes to test on CUB. We follow similar settings for testing on Flowers 102.
Even in these challenging circumstances our method offers improvement over the baselines(\S~\ref{sec:domaindiff}).

For CUB, NABirds and iNaturalist we also have taxonomy information including family, order and scientific name. We also append separate texts containing these to the category-wise text corpus to show improvements  (Tab.~\ref{tab:othertexts}).

\textbf{NeWT}~\cite{van2021benchmarking} provides a benchmark for a set of 164 complex binary classification tasks in the natural world that extend beyond species classification. These tasks include determining 1) appearance 2) behavior, 3) context, 4) counting and 5) gestalt. NeWT contains 36k images with 200-400 images per task. We randomly select 50 of the 164 tasks to evaluate our trained model. 
We manually associate two texts for each task, positive and negative. For example, ``a photo of a raptor bird which is not on a utility pole" and ``a photo of a raptor bird which is on a utility pole". We show improvements over CLIP (\S~\ref{sec:noveltasks}). All details of texts used as well as categories selected are in the Appendix.

\subsection{Implementation Details}
\label{sec:implement}
For \textbf{generating category-level texts} for training, we utilize the ``gpt-4-0613" API. We set the temperature parameter as 0 so that texts generated are deterministic. 




For all queries concerning the classes of iNaturalist (both visual and location) we also append the type of organism as well as it's scientific name in the question. 
For example, for the class ``Bay Laurel" the query for location information is 


\begin{quote}
    \makebox[\linewidth]{%
        \colorbox{lightblue}{%
            \hspace*{0mm} 
            \begin{minipage}{\dimexpr\linewidth+10\fboxsep\relax} 
                \fontsize{9pt}{10pt}\selectfont 
                Where can we find a Bay Laurel, a type of plant with scientific name Laurus nobilis? Produce a list of habitat and geographic location information that can be used to identify the plant.
            \end{minipage}%
        }%
    }
\end{quote}

This is required because there exist organisms with the same common name but different domains. Also, we need to append the scientific name as otherwise GPT4 does not recognise the organism in many cases. We provide more details in Appendix.

Additionally, we experiment with using \textbf{taxonomy information} for training and testing on datasets where it is available (Tab.~\ref{tab:othertexts}). We form the following texts


\begin{quote}
    \makebox[\linewidth]{%
        \colorbox{lightblue}{%
            \hspace*{-3mm}
            \begin{minipage}{\dimexpr\linewidth+12\fboxsep\relax}
                \fontsize{8pt}{10pt}\selectfont
                \begin{enumerate}[nosep]
                    \item a photo of \texttt{[class]} \texttt{[domain]}, with scientific name \texttt{[s\_{name}]}
                    \item a photo of \texttt{[class]} \texttt{[domain]}, with family name \texttt{[family]}
                    \item a photo of \texttt{[class]} \texttt{[domain]}, of the order \texttt{[order]}
                \end{enumerate}
            \end{minipage}%
        }%
    }
\end{quote}

While \textbf{fine-tuning} using the texts obtained from an LLM, we train for only 15 epochs on each dataset. On iNaturalist, we train for only 5 epochs. We find hyperparameters by splitting the train classes into two equal parts, training on the first half, and validating on the second. We then fix the best hyperparameters found and train on all train classes.


The CLIP architecture consists of an image encoder and a text encoder. Both contain transformers followed by a linear projection layer at the end. We use different learning rate and weight decay for the projection layers compared to rest of the encoders. The temperature parameter $\tau$ in our model (\S~\ref{sec:finetune}) is trainable. We provide the details of the initialization of $\tau$, the momentum parameter for the EMA encoder as well as learning rates and weight decays of every parameter for all datasets in the Appendix.

\subsection{Baselines}
\label{sec:baselines}
In this section, we discuss the various methods for which we compare zero-shot classification accuracy.

\noindent \textbf{CLIP} refers to pre-trained CLIP tested with ``a photo of a \texttt{[class]} \texttt{[domain]}" texts like the original paper.

\noindent \textbf{CLIP + A} is evaluating pre-trained CLIP with attributes obtained from LLMs as outlined in \S~\ref{sec:eval}.

\noindent \textbf{CLIP\textsuperscript{FT}} involves fine-tuning CLIP on training classes using ``a photo of \texttt{[class]} \texttt{[domain]}" texts and evaluating on test classes using ``a photo of \texttt{[class]} \texttt{[domain]}" texts.

\noindent \textbf{CLIP\textsuperscript{FT} + A} is our method where we fine-tune CLIP~(\S~\ref{sec:finetune}) using attributes obtained from a LLM~(\S~\ref{sec:queryllm}) for training classes and evaluate using LLM attributes of testing classes at test time as described in \S~\ref{sec:eval}.

\noindent \textbf{CLIP-A-self~\cite{maniparambil2023enhancing}} is prior work which uses text obtained from GPT to train an adapter network attached after the text and image encoders of CLIP. For comparing with this, we use the numbers stated by them under their training and evaluation scheme. We test our model on the same classes as them to show improvement.

\vspace{-1mm}
\section{Results}
In this section, we compare our method to baselines and evaluate it under various settings. We discuss our performance improvements over various datasets and architectures. We show that for natural domains using taxonomy and habitat information offers improvements with habitat information especially being a strong factor. Our model scales across architectures and needs only a few epochs of training. We further show that our method performs better than baseline CLIP even under more difficult evaluation settings such as 1) using texts from different LLM models during testing and training; 2) training a model in a domain very different from testing domain, and 3) evaluating on tasks other than identifying categories at test time. Additionally, we discuss other training strategies for category-level fine-tuning and how they perform.

\subsection{Comparison with Baselines}
\label{sec:comparepriorart}
Tab.~\ref{tab:maintable} compares our method to three baselines CLIP, CLIP + A and CLIP\textsuperscript{FT}, all evaluated on \textbf{unseen classes}. Here we use the ViT B/32 architecture for all methods. Our method offers considerable improvements over pre-trained CLIP when using ``a photo of a \texttt{[class]} \texttt{[domain]}" text and when using GPT generated text. In difficult fine-grained domains such as CUB, Stanford Cars and FGVC Aircrafts pre-trained CLIP does not utilize text attributes generated by GPT resulting in negligible improvement (decrease on Stanford Cars) compared to ``a photo of a \texttt{[class]} \texttt{[domain]}" (see CLIP + A vs CLIP). This motivates the need to fine-tune using these attributes, resulting in significant improvement across all datasets. We also compare to fine-tuning CLIP with ``a photo of a \texttt{[class]} \texttt{[domain]}" texts (CLIP\textsuperscript{FT}) to show that the improvement our method achieves is not due to just being trained on images of concerned domain. 


We compare our method to previous work CLIP-A-self~\cite{maniparambil2023enhancing} in Tab.~\ref{tab:compare}. We follow~\cite{maniparambil2023enhancing} and use the ViT B/16 architecture and only 16 images per class for training. Again we evaluate on \textbf{unseen classes}. Our method outperforms CLIP-A-self significantly across all datasets. Also, our method offers substantial improvement over CLIP, showing that it \textbf{scales across architectures}. We discuss why CLIP-A-self underperforms in detail in the Appendix. We also show results on the \textbf{14 datasets} benchmark in Appendix.

\subsection{Using more than just Visual Information}
\label{sec:vistaxhab}
We explore using information other than visual attributes for natural domains such as birds and flowers. For humans identifying a bird in an image it is crucial to know where the image was taken, because that reveals habitat and location information. We therefore query GPT for an organism's habitat and geographic range. In Tab.~\ref{tab:othertexts} we show that for CUB using \textbf{only habitat information performs better than using only visual information}. A reason for this is that habitat information describes the background of the images of birds, which is helping to differentiate between categories. 
We also show that combining visual + taxonomy + habitat information for CUB and visual + habitat information for Flowers102 offers best improvement. 

\subsection{Training on External Domains}
\label{sec:domaindiff}
We now evaluate under more difficult settings. We train and test on different datasets, always removing any overlapping classes. For training on iNat and NABirds we use all visual + taxonomy + habitat information. While testing on CUB we use visual + taxonomy + habitat. For Flowers 102 we use visual + habitat information. Tab.~\ref{tab:crossdomain} we show the accuracy on CUB test classes improves considerably when training using NABirds and iNat even though the images of these datasets have a distribution shift \wrt CUB. More strikingly, we show that even when we remove all bird classes from iNat we still offer improvement on CUB test classes compared to CLIP + A (52.83 $\rightarrow$ 53.89). Similarly when we remove all plant classes from iNat, we still get improvement on Flowers 102 test set. This proves that our \textbf{model is also able to generalize well}. It is learning to associate fine-grained attributes to images irrespective of the domain differences in training and testing.

\subsection{Using Novel Texts during Test Time}
\label{sec:noveltexts}
We evaluate how our model would perform in the absence of the LLM used to generate training texts, during test time (Tab~\ref{tab:otherllm}). 
We use GPT3.5 turbo (0613) and LLaMA2-7B~\cite{touvron2023LLaMA} for generating visual and habitat texts. We show that our method \textbf{consistently improves performance over pre-trained CLIP for all types of texts} explored. \footnote{LLaMA model does not always produce texts in the specified format and thus needs post-processing. The texts formed finally are considerably different grammatically from the sentences our model has been trained on.} Please refer to Appendix for examples of texts produced. Our model also improves performance over pre-trained CLIP while using ``a photo of a \texttt{[class]} \texttt{[domain]}" texts.

Reed \etal~\cite{reed2016learning} present a dataset of human-labelled captions per image for the CUB dataset. We use these \textbf{ground truth texts at test time} to evaluate our model. The dataset contains 10 captions for every image, which are all visual attributes of the bird in the image. 
Since many attributes in the captions are repeated for each image as well as across images of the same class, and to limit the size of the text corpus, we randomly select one caption per image of a given class and aggregate them to form a category corpus. 
In Tab.~\ref{tab:otherllm} last row, we notice that CLIP does better using these image-level GT texts compared to using GPT4 Visual texts which were category-level (row 2). However, our method still outperforms showing that it is able learn meaningful attributes through noisy labels.

\subsection{Evaluation on Novel Tasks}
\label{sec:noveltasks}
We use the NeWT~\cite{van2021benchmarking} benchmark to evaluate on tasks beyond categorization. These include identifying age, attribute, health, photo quality, species, context and behavior. We evaluate the model and baseline CLIP using average Mean Average Precision (MAP) across tasks. Our model trained on iNaturalist using visual + taxonomy + location information outperforms baseline CLIP: 60.25 vs 61.90 MAP $\rightarrow$ \textbf{4.1\% relative error reduction}. We present all tasks and texts we use to evaluate as well as MAP per task in the Appendix. Below is an example prediction:

\begin{figure}[ht]
  \centering
\includegraphics[width=\columnwidth]{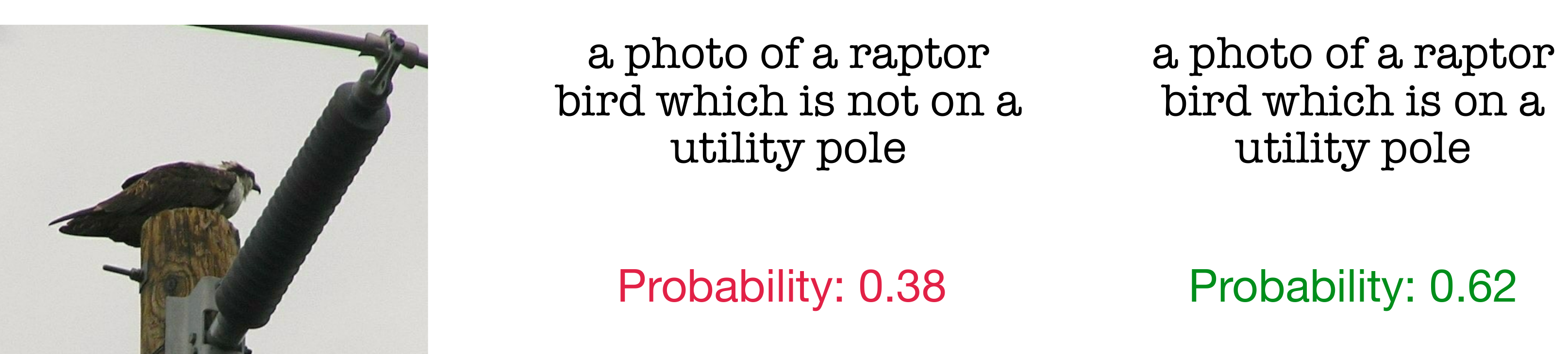}
\end{figure}
\subsection{Resource Requirements}
We fine-tune our model using a 1024 batch size on a single NVIDIA A100 80GB. We need to train only for a maximum of 15 epochs. For smaller datasets such as CUB, FGVC Aircraft, and Stanford Cars, this takes less than 5 minutes. For iNaturalist, which is a large dataset, we train only for 5 epochs, taking about 4 hours. 

The cost of using the GPT-4 API to query text descriptions for a dataset with 100-200 classes (such as CUB) is about \$1-\$5. This is low because it scales with the number of categories and not the number of images. Generating captions per image is both time-consuming and expensive; an estimate for doing this for the CUB dataset using the GPT-4 Vision API is more than \$100. This cost scales significantly for larger datasets with more images.

\subsection{Alternate Training Strategies}
\label{sec:alttrain}

Here we discuss other training strategies for improving VLMs using category-level training data. Our \textbf{\textit{simple fine-tuning strategy}} of stochastically pairing images with texts within categories \textbf{is simple, efficient, and offers similar improvements} compared to more complex approaches.

Firstly, since we pair a given image of a category to every text of that category, we might be pairing texts that describe attributes that are not visible. For CUB dataset we have the \textbf{ground truth visibility annotations} of various bird parts, which we use to ignore texts that are occluded for each image. This strategy offers improvement to our scores (54.23 $\rightarrow$ 54.47). However, this depends on  visibility information that is time-consuming to generate. 

The next step is to assume that pre-trained CLIP itself is able to correctly identify if a part is visible. 
Assuming this we \textbf{mask texts during training time based on CLIP predictions} by doing 1) a forward pass for a image and all texts of the category to find the texts above a threshold that can be paired with the image, 2) max pooling at instance level for images and texts.
We find that none of these strategies offer any improvement and that pairing images with low scoring texts also (like in our method) is improving performance because CLIP does not accurately identify which fine-grained attributes correctly correspond to given image. 

We also try well-known semi-supervised learning strategies such as FixMatch~\cite{sohn2020fixmatch} and knowledge distillation~\cite{chen2020big}.
We find that these offer small ($<$ 0.2$\%$) to no improvements over our method. Please see the Appendix for details of implementation of all the methods and accuracies.

\subsection{Performance of image captioning models.} We test the recently released GPT4 Vision API for checking quality of image captions obtained. Even though it performs better than previous captioning models such as LLaVA~\cite{liu2023visual} and BLIP2~\cite{li2022blip}, we find that the captions obtained are general descriptions without fine-grained details. The captions are specific to the image but do not describe information helpful to identify the category. An example is:


\begin{figure}[ht]
  \centering
  \begin{minipage}[t]{0.18\linewidth}
    \includegraphics[width=\linewidth]{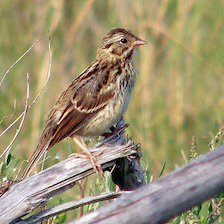}
    \label{fig:yourlabel}
  \end{minipage}
  \hfill 
  \begin{minipage}[t]{0.8\linewidth} 
    \raisebox{\dimexpr\topskip+0.36\height}{
      \colorbox{lightblue}{%
        \parbox{\dimexpr\linewidth-2\fboxsep}{
          \fontsize{9pt}{10pt}\selectfont 
          A slender, streaked brown songbird with keen eyes and a pointed beak perches atop a weathered wooden fence post amidst a backdrop of natural grassland under a clear blue sky.
        }%
      }%
    }
  \end{minipage}
\end{figure}

\vspace{-3mm}


\noindent For this image of a Vesper Sparrow, GPT4 provides a general description of the bird and suggests the presence of clear blue sky which is not visible in the image. We provide more detail including prompt and examples in Appendix.

\section{Limitation}
Since our method is trained on texts generated by LLMs, it is important to verify the correctness of these texts to assess the level of noise in our training dataset. As described in \S~\ref{sec:queryllm}, we conduct spot checks across some categories on our evaluation sets. However, for our larger training datasets, vetting becomes impractical. Improved performance on human-generated texts on CUB, as well as the vetted evaluation sets, supports our model's ability to learn meaningful information from somewhat noisy training data.

\section{Conclusion}
We present a method to improve the zero-shot performance of VLMs using attributes generated by LLMs on fine-grained domains. Our evaluation strategy involves testing the trained model on unseen classes, texts generated from different LLMs as well as humans, dissimilar domains, and novel tasks. We show that fine-tuning CLIP using category-level descriptions from GPT4 significantly improves performance compared to baselines in this challenging downstream evaluation framework. Our findings suggests that habitat and geographic priors are equally effective and complementary to visual information for zero-shot classification in natural domains. We publicly release our benchmark across all 14 datasets.

\section{Acknowledgements}
We thank Aaron Sun, Gustavo Perez, Rangel Daroya, and Mustafa Chasmai for participating in the verification of attributes generated by GPT4. The project is supported in part by NSF Grants \#2329927 and \#1749833. Our experiments were performed on the GPU cluster funded by the Mass. Technology Collaborative.


{\small
\bibliographystyle{ieee_fullname}
\bibliography{egbib}
}

\clearpage
\appendix

\onecolumn
\vspace{-2mm}
\noindent{\Large \textbf{Appendix}}
\vspace{-2mm}
\section{Category-level Texts produced by LLMs}
We provide more examples of texts produced for different prompts, including both visual and habitat. For prompting about visual attributes we set `System Role' as ``You are a helpful assistant who can identify any \texttt{[domain]} based on its photo." For prompting about habitat attributes we set `System Role' as ``You are a helpful assistant who knows about the habitat of any \texttt{[domain]}."
\subsection{GPT4 texts used in our method}
\vspace{-2mm}

\begin{figure}[ht]
\begin{minipage}{0.9\textwidth}
\resizebox{\textwidth}{!}{
\begin{tabular}{ll}
\textbf{Prompt} \\
\begin{tabular}[c]{@{}l@{}}What characteristics can be used to differentiate a \textit{White Spruce},  \\ a type of plant with scientific name Picea glauca from other organisms based on just a photo?\\ Produce an exhaustive list of all attributes or any information that can be used to identify the plant uniquely.\\ Texts should be of the form ``Plant with \textless{}description\textgreater{}". \\ Ensure to structure your response as a list of single sentences.\end{tabular} \\
\textbf{Output} &  \\
\multicolumn{2}{l}{\begin{tabular}[c]{@{}l@{}}- Plant with short, blunt, and four-sided needles that are blue-green to green in color.\\- Plant with needles that are spirally arranged on the branches.\\- Plant with a conical shape, often with a straight trunk.\\- Plant with bark that is thin, scaly, and grayish-brown in color.\\- Plant with cones that are slender, cylindrical, and hang down from the branches.\\- Plant with cones that are light brown and have thin, flexible scales.\\- Plant with branches that are densely covered with needles, giving it a full appearance.\\- Plant with a height that can reach up to 60 feet or more.\\- Plant with a habitat often found in cold, northern climates, such as the boreal forests of Canada and Alaska.\\- Plant with a preference for well-drained, sandy or gravelly soils.\\- Plant with a growth pattern that is slow to moderate.\\- Plant with a crown that is narrow and pointed.\\- Plant with lower branches that often droop downwards.\\- Plant with a root system that is shallow and wide-spreading.\\- Plant with a tolerance for cold temperatures and harsh winter conditions.\\- Plant with a tendency to grow in pure stands or mixed with other conifer species.\\- Plant with a lifespan that can exceed 200 years.\\- Plant with a resistance to pests and diseases, making it a hardy species.\\- Plant with a usage in landscaping for windbreaks, privacy screens, or as a specimen tree.\\
- Plant with a significance in the timber industry for its wood, which is used for pulp and construction.\end{tabular}} \\
 &  \\
\textbf{Prompt}\\
\begin{tabular}[c]{@{}l@{}}Where can we find a \textit{White Spruce},  a type of plant with scientific name Picea glauca?\\ Produce a list of all habitat and geographic information that can be used to identify the plant.\\ Each text produced should start with ``Plant which". \\ Ensure to structure your response as a list of single sentences.\end{tabular} \\
\textbf{Output} & \\
\multicolumn{2}{l}{\begin{tabular}[c]{@{}l@{}}- Plant which thrives in the cold, boreal forests of North America.\\- Plant which is native to the northern parts of North America, including Alaska and Canada.\\- Plant which can be found growing in the high altitudes of the Rocky Mountains.\\- Plant which is commonly found in the taiga and the boreal forests of North America.\\- Plant which can survive in a variety of soils, including sandy, loamy, and clay soils.\\- Plant which can be found in the northeastern and northwestern parts of the United States.\\- Plant which is often found in cool, moist locations, such as near rivers and lakes.\\- Plant which can tolerate harsh winter conditions and is often found in areas with heavy snowfall.\\- Plant which is a dominant species in the North American boreal forest, also known as the taiga.\\- Plant which can be found in the northern parts of the United States, including Maine, Wisconsin, and Minnesota.\end{tabular}}
\end{tabular}}
\end{minipage}%
\begin{minipage}{0.1\textwidth}
\includegraphics[width=\columnwidth]{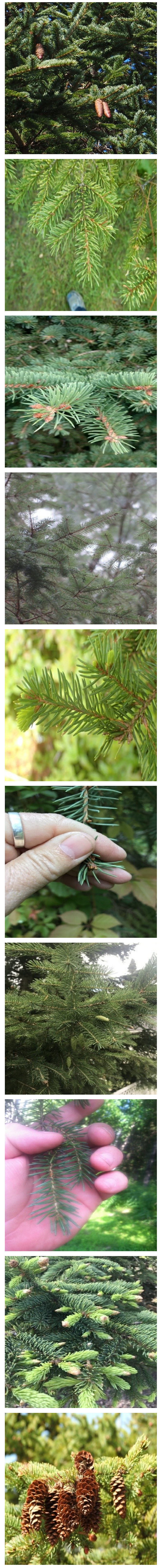}
\end{minipage}%
\caption{Example of texts produced for a category ``White Spruce" of the iNaturalist dataset using GPT4-0613, example images on the right.}
\end{figure}

\begin{figure}[ht]
\begin{minipage}{0.9\textwidth}
\resizebox{\textwidth}{!}{
\begin{tabular}{ll}
\textbf{Prompt}\\
\begin{tabular}[c]{@{}l@{}}What characteristics can be used to differentiate a \textit{Common Tern} from other birds based on just a photo?\\ Produce an exhaustive list of all attributes or any information that can be used to identify the bird uniquely. \\ Texts should be of the form ``a bird with \textless{}characteristic feature\textgreater{}". \\ Ensure to structure your response as a list of single sentences.\end{tabular}\\
\textbf{Output} &  \\
\multicolumn{2}{l}{\begin{tabular}[c]{@{}l@{}}- A bird with a slender, pointed bill that is usually orange with a black tip.\\- A bird with a black cap on its head that extends to the eyes.\\- A bird with a deeply forked tail, which gives it a distinctive 'swallow-like' appearance.\\- A bird with long, narrow wings that are grey on top and lighter underneath.\\- A bird with short, red legs.\\- A bird with a white underbelly and chest.\\- A bird with a size of approximately 31-35 cm in length and a wingspan of 70-80 cm.\\- A bird with a distinctive hovering flight pattern before diving to catch fish.\\- A bird with a sharp, piercing call that sounds like 'kee-arr'.\\- A bird that is often seen in coastal areas, particularly during the breeding season.\\- A bird that is often seen in large flocks, particularly during migration.\\- A bird that is commonly found in the northern parts of North America, Europe, and Asia during the breeding season.\\- A bird that migrates long distances, often to the southern parts of South America and Africa during the non-breeding season.\\- A bird that is often seen diving into the water to catch small fish and invertebrates.\\- A bird with a sleek, streamlined body that is well adapted for its aquatic lifestyle.\\- A bird that is often seen perched on buoys, posts, and other structures near water.\\- A bird that is known for its aggressive behavior towards intruders, including humans, during the breeding season.\\- A bird that lays its eggs on the ground in a shallow scrape, often in a colony with other terns.\\- A bird with juveniles that have a scaly pattern on their wings and back.\end{tabular}} \\
 &  \\
\textbf{Prompt}\\
\begin{tabular}[c]{@{}l@{}}Where can we find a \textit{Common Tern}?\\ Produce a list of habitat and geographic location information that can be used to identify the bird.\\ Each text produced should start with ``a bird". \\ Ensure to structure your response as a list of single sentences.\end{tabular} \\
\textbf{Output} &  \\
\multicolumn{2}{l}{\begin{tabular}[c]{@{}l@{}}- A bird which can be found in coastal areas and inland water bodies across North America, Europe, and Asia\\ during the breeding season.\\- A bird which migrates to warmer coastal areas, offshore islands, and open seas in the tropics and Southern\\ Hemisphere for the winter.\\- A bird which nests in colonies on the ground in areas with loose sand, gravel, shell, or cobble pebbles typically\\ less than 350 feet from the water.\\- A bird which can be seen in estuaries, salt marshes, and sandy or rocky shorelines during migration and winter.\\- A bird which can also be found in man-made habitats such as dredge spoil islands and gravel rooftops.\\- A bird which is often seen hovering over water before diving down to catch a fish.\\- A bird which is known for its long-distance migrations, with some individuals traveling from the Arctic to the \\ Antarctic and back each year.\\- A bird which is easily identifiable by its slender body, long pointed wings, forked tail, and black cap on its head.\end{tabular}}
\end{tabular}}
\end{minipage}%
\begin{minipage}{0.1\textwidth}
\includegraphics[width=\columnwidth]{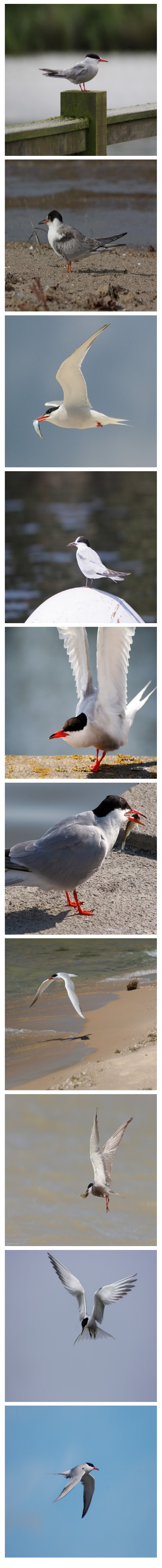}
\end{minipage}%
\caption{Example of texts produced for a category ``Common Tern" of the CUB dataset using GPT4-0613, example images on the right.}
\end{figure}

\clearpage
\newpage

\subsection{LLaMA texts used during evaluation}

\begin{table}[ht]
\begin{tabular}{ll}
\textbf{Prompt}\\
\begin{tabular}[c]{@{}l@{}}What characteristics can be used to differentiate a \textit{Common Tern} from other birds based on just a photo?\\ Produce an exhaustive list of all attributes or any information that can be used to identify the bird uniquely. \\ Texts should be of the form ``a bird with \textless{}characteristic feature\textgreater{}". \\ Ensure to structure your response as a list of single sentences.\end{tabular}\\
\textbf{Output} &  \\
\multicolumn{2}{l}{\begin{tabular}[c]{@{}l@{}}- A bird with a white forehead and a distinctive black stripe running across its eyes.\\
- A bird with a pale yellow bill that is longer than its head.\\
- A bird with a dark grey back and wings, and a light grey underside.\\
- A bird with a distinctive black stripe running along its tail feathers.\\
- A bird with a distinctive black band on its outermost tail feathers.\\
- A bird with a pale yellow patch on its forehead that is shaped like a crescent moon.\\
- A bird with a distinctive black stripe running along its neck.\\
- A bird with a dark grey crown and a light grey nape.\\
- A bird with a pale yellow patch on its upperwing that is shaped like a crescent moon.\\
- A bird with a distinctive black stripe running along its upperwing.\\
- A bird with a dark grey patch on its upperwing that is shaped like a triangle.\\
- A bird with a pale yellow patch on its lowerwing that is shaped like a crescent moon.\\
- A bird with a distinctive black stripe running along its lowerwing.\\
- A bird with a dark grey patch on its lowerwing that is shaped like a triangle.\\
- A bird with a distinctive black stripe running along its legs and feet.\\
- A bird with a pale yellow patch on its upperwing that is surrounded by a dark grey border.\\
- A bird with a distinctive black stripe running along its upperwing that is surrounded by a pale yellow border.\\
- A bird with a pale yellow patch on its lowerwing that is surrounded by a dark grey border.\\
- A bird with a distinctive black stripe running along its lowerwing that is surrounded by a pale yellow border.\end{tabular}} \\
 &  \\
\textbf{Prompt}\\
\begin{tabular}[c]{@{}l@{}}Where can we find a \textit{Common Tern}?\\ Produce a list of habitat and geographic location information that can be used to identify the bird.\\ Each text produced should start with ``a bird". \\ Ensure to structure your response as a list of single sentences.\end{tabular} \\
\textbf{Output} &  \\
\multicolumn{2}{l}{\begin{tabular}[c]{@{}l@{}}- A bird of freshwater lakes and rivers, the Common Tern can also be found in inland areas with large bodies of water,\\ such as reservoirs, ponds, and wetlands. In North America, they can be spotted in the Great Lakes region and in the central\\ United States. In Europe, they can be found in the British Isles, France, and Germany. In Asia, they can be seen in China,\\ Korea, and Japan. In Australia, they can be found in the southeastern region.\\
- A bird of the Arctic tundra, the Common Tern can be found in the northernmost parts of Europe, Asia, and North America,\\ including Alaska and Canada. In these regions, they can be seen nesting on rocky cliffs and in coastal areas.\end{tabular}}
\end{tabular}
\caption{Example of texts produced for category ``Common Tern" of the CUB dataset using LLaMA2-7B.}
\end{table}

\clearpage
\section{LLMs for Fine-grained Image Captioning}
We present captions generated using \href{https://replicate.com/gfodor/instructblip/versions/ca869b56b2a3b1cdf591c353deb3fa1a94b9c35fde477ef6ca1d248af56f9c84}{InstructBLIP}, \href{https://huggingface.co/spaces/Vision-CAIR/minigpt4}{MiniGPT4} and recently released OpenAI's GPT4 Vision API. We specifically ask for concise single sentences in less than 70 words such that can be used to train CLIP using image-level supervision. This is in consideration with the fact that CLIP has a token length of 77. We use the following prompt:

\begin{quote}
\centering
    \makebox[\linewidth]{%
        \colorbox{lightblue}{%
            \hspace*{0mm} 
            \begin{minipage}{\dimexpr\linewidth+10\fboxsep\relax} 
            \centering
                \fontsize{10pt}{11pt}\selectfont 
                Describe this image in fine-grained detail with as much information as possible in a single sentence less than 70 words.
            \end{minipage}%
        }%
    }
\end{quote}

Below we show examples of texts generated.

\begin{figure}[ht]
  \centering
\includegraphics[width=\columnwidth]{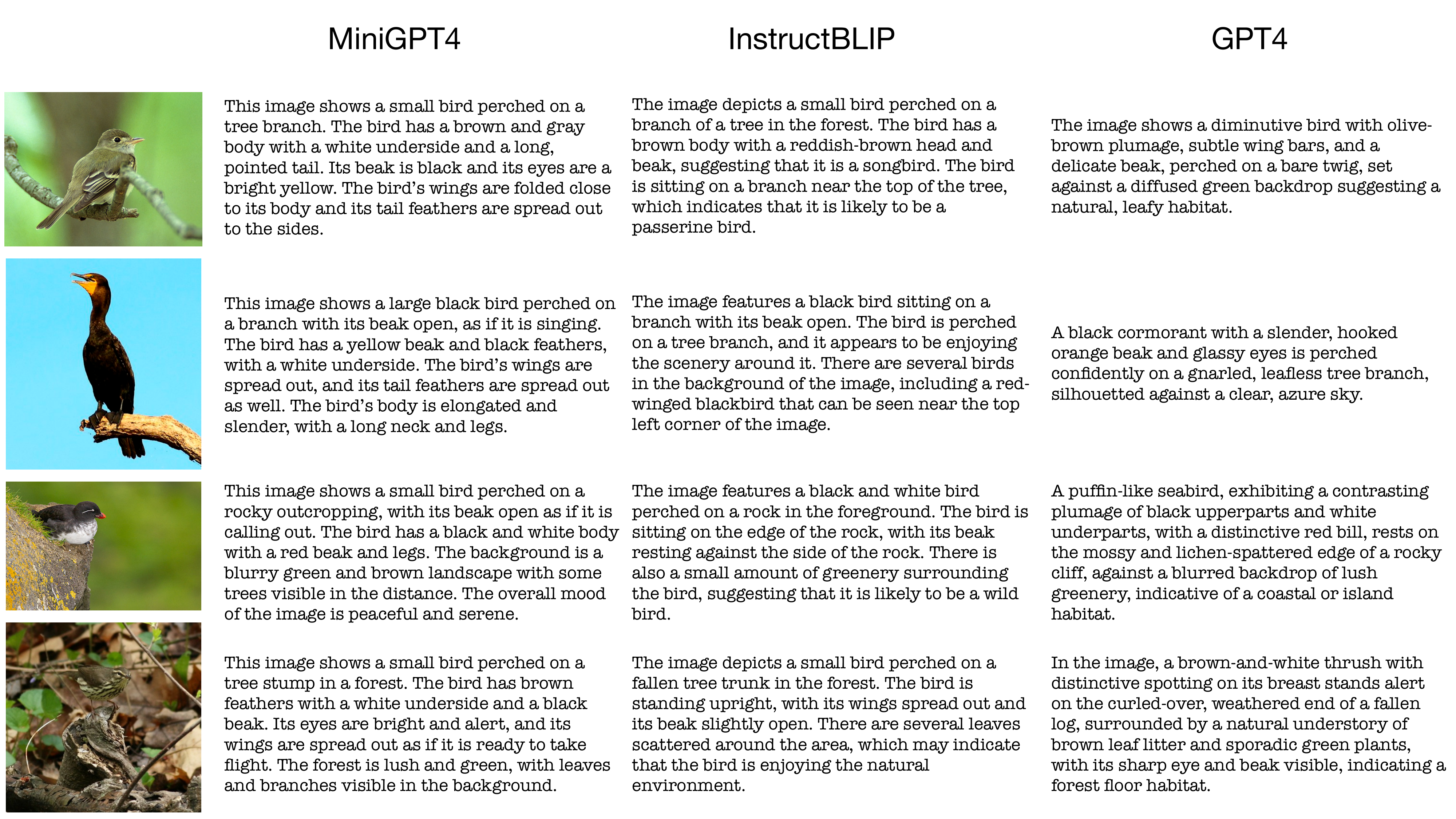}
\end{figure}

MiniGPT4 and InstructBLIP are not able to produce a single long sentence, rather a set of sentences. We show all those sentences in the figure such that context length is less than 77. \textbf{GPT4 is much more accurate} than MiniGPT4 and InstructBLIP, but still makes some errors such as in the color of the bird in the first and last image. It identifies the color of the upperparts of the first bird - an Acadian Flycatcher - as olive brown whereas it is olive green. In the last bird it identifies the bird as brown-and-white whereas it is brown and yellow. It should be noted that our \textbf{category level descriptions capture the correct colors}. The descriptions are also very coarse descriptions of few body parts.

\section{Training Hyperparameters}
We list the training hyperparameters for all datasets. The temperature parameter is learned using learning rate (lr) 1E-2 and weight decay (wd) 1E-6 for all. The momentum parameter is set to 0.98 for all.

\begin{table}[ht]
\resizebox{\textwidth}{!}{
\begin{tabular}{lrrrrrrlrrrrrrrr}
 & \multicolumn{1}{c}{CUB} & \multicolumn{1}{c}{FGVC Aircraft} & \multicolumn{1}{c}{Stanford Cars} & \multicolumn{1}{c}{Flowers102} & \multicolumn{1}{c}{NABirds} & \multicolumn{2}{c}{iNat} & \multicolumn{1}{c}{EuroSAT} & \multicolumn{1}{c}{Food101} & \multicolumn{1}{c}{ImageNet} & \multicolumn{1}{c}{CalTech101} & \multicolumn{1}{c}{DTD} & \multicolumn{1}{c}{Oxford Pets} & \multicolumn{1}{c}{Sun397} & \multicolumn{1}{c}{UCF101} \\ \hline
lr proj & 6E-07 & 4E-07 & 3E-07 & 7E-07 & 8E-07 & \multicolumn{2}{r}{1E-07} & 2E-06 & 1E-07 & 2E-06 & 1E-07 & 5E-07 & 8E-07 & 1E-06 & 8E-07 \\
lr main & 1E-07 & 1E-07 & 1E-07 & 1E-07 & 1E-07 & \multicolumn{2}{r}{5E-08} & 5E-07 & 5E-08 & 5E-07 & 5E-08 & 2E-07 & 2E-07 & 5E-07 & 2E-07 \\
wd proj & 1E-01 & 1E-06 & 1E-02 & 1E-02 & 1E-06 & \multicolumn{2}{r}{1E-03} & 1E-06 & 1E-06 & 1E-06 & 1E-03 & 1E-06 & 1E-04 & 1E-06 & 1E-06 \\
wd main & 1E-01 & 1E-06 & 1E-02 & 1E-03 & 1E-06 & \multicolumn{2}{r}{1E-03} & 1E-06 & 1E-06 & 1E-06 & 1E-03 & 1E-06 & 1E-04 & 1E-06 & 1E-06 \\
temperature init & 1.3 & 1.8 & 1.6 & 2 & 1 & \multicolumn{2}{r}{1} & 1.8 & 2 & 1.8 & 2 & 2 & 2 & 2 & 2
\end{tabular}}
\end{table}

\clearpage
\section{Texts and Per-task MAP on NeWT}
\begin{table}[ht]
\resizebox{\textwidth}{!}{
\begin{tabular}{lllcc}
\textbf{Task} & \textbf{Negative Text} & \textbf{Positive Text} & \textbf{CLIP} & \textbf{Ours} \\ \hline
inat\_non\_species\_birds\_near\_signs & a photo of a bird which is not on a road sign & a photo of a bird which is on a road sign & 95.75 & 97.94 \\
inat\_non\_species\_diseased\_zebra\_finch & a photo of a zebra finch bird which is healthy & a photo of a zebra finch bird which is diseased & 61.10 & 61.20 \\
inat\_non\_species\_intersex\_mallards & a photo of a mallard bird which is not intersex & a photo of a mallard bird which is intersex & 48.12 & 47.72 \\
inat\_non\_species\_mating\_aligator\_lizard & a photo of a aligator lizard mating & a photo of a aligator lizard not mating & 34.66 & 34.69 \\
inat\_non\_species\_mating\_danaus\_plexippus & a photo of danaus plexippus which are mating & a photo of danaus plexippus not mating & 64.20 & 69.40 \\
inat\_non\_species\_mating\_toxomerus\_marginatus & a photo of a toxomerus marginatus mating & a photo of a toxomerus marginatus not mating & 37.84 & 38.22 \\
inat\_non\_species\_white\_american\_robin & a photo of a white robin bird & a photo of an american robin bird & 36.94 & 36.72 \\
inat\_observed\_Allegheny\_Mountain\_Dusky\_Salamander\_vs\_Dusky\_Salamander & a photo of a Allegheny Mountain Dusky Salamander & a photo of a Dusky Salamander & 46.30 & 50.10 \\
inat\_observed\_Belize\_Crocodile\_vs\_American\_Crocodile & a photo of a Belize Crocodile & a photo of an American Crocodile & 62.22 & 64.30 \\
inat\_observed\_California\_Sea\_Lion\_vs\_Steller\_Sea\_Lion & a photo of a california sea lion & a photo of a stellar sea lion & 58.22 & 63.28 \\
inat\_observed\_Common\_Grass\_Yellow\_vs\_Three-spotted\_Grass\_Yellow & a photo of a Common Grass Yellow & a photo of a Three-spotted Grass Yellow & 47.78 & 48.03 \\
inat\_observed\_Eastern\_Oyster\_vs\_Pacific\_Oyster & a photo of a Eastern Oyster & a photo of a Pacific Oyster & 63.34 & 72.70 \\
inat\_observed\_Flea\_Jumper\_vs\_Asiatic\_Wall\_Jumping\_Spider & a photo of a Flea Jumper & a photo of a Asiatic Wall Jumping Spider & 50.00 & 51.10 \\
inat\_observed\_Jelly\_Ear\_vs\_Ear\_fungus & a photo of a Jelly Ear & a photo of a Ear fungus & 58.16 & 59.22 \\
inat\_observed\_Northern\_Cinnabar\_Polypore\_vs\_Cinnabar\_Bracket & a photo of a Northern Cinnabar Polypore & a photo of a Cinnabar Bracket & 49.53 & 52.72 \\
inat\_observed\_Rough\_Green\_Snake\_vs\_Smooth\_Greensnake & a photo of a Rough Green Snake & a photo of a Smooth Greensnake & 52.97 & 57.34 \\
inat\_observed\_Southern\_Black\_Widow\_vs\_Western\_Black\_Widow & a photo of a Southern Black Widow & a photo of a Western Black Widow & 48.44 & 47.72 \\
inat\_observed\_Western\_Grey\_Kangaroo\_vs\_Eastern\_Grey\_Kangaroo & a photo of a Western Grey Kangaroo & a photo of a Eastern Grey Kangaroo & 60.70 & 62.56 \\
inat\_observed\_southern\_cattail\_vs\_lesser\_reedmace & a photo of a southern cattail & a photo of a lesser reedmace & 52.75 & 52.60 \\
inat\_unobserved\_amanita\_flavorubens\_v\_amanita\_xanthocephala & a photo of a amanita flavorubens & a photo of a amanita xanthocephala & 52.72 & 47.80 \\
inat\_unobserved\_armillaria\_luteobubalina\_v\_armillaria\_novae-zelandiae & a photo of a Armillaria luteobubalina & a photo of a Armillaria novae-zelandiae & 35.20 & 35.72 \\
inat\_unobserved\_chloris\_verticillata\_v\_chloris\_cucullata & a photo of a chloris verticillata & a photo of a chloris cucullata & 39.68 & 41.56 \\
inat\_unobserved\_cladonia\_squamosa\_v\_cladonia\_portentosa & a photo of a Cladonia squamosa & a photo of a Cladonia portentosa & 52.20 & 61.22 \\
inat\_unobserved\_cuphea\_aequipetala\_v\_cuphea\_hyssopifolia & a photo of a Cuphea aequipetala & a photo of a Cuphea hyssopifolia & 43.90 & 49.97 \\
inat\_unobserved\_pinus\_clausa\_v\_pinus\_mugo & a photo of a Pinus clausa & a photo of a Pinus mugo & 71.00 & 66.06 \\
inat\_unobserved\_podarcis\_virescens\_v\_podarcis\_guadarramae & a photo of a Podarcis virescens & a photo of a Podarcis guadarramae & 47.29 & 46.80 \\
inat\_unobserved\_turdus\_torquatus\_v\_turdus\_atrogularis & a photo of a Turdus torquatus & a photo of a Turdus atrogularis & 70.30 & 79.80 \\
ml\_age\_black\_bellied\_plover & a photo of a black bellied plover bird which is not an adult & a photo of a black bellied plover bird which is an adult & 45.96 & 41.12 \\
ml\_age\_coopers\_hawk & a photo of a cooper's hawk bird which is not an adult & a photo of a cooper's hawk bird which is an adult & 53.47 & 68.90 \\
ml\_age\_sanderling & a photo of a sanderling which is not an adult & a photo of a sanderling which is an adult & 54.06 & 53.84 \\
ml\_bio\_is\_at\_flower & a photo of a bird which is not at a flower & a photo of a bird which is at a flower & 94.30 & 94.70 \\
ml\_bio\_raptor\_utility\_pole & a photo of a raptor bird which is not on a utility pole & a photo of a raptor bird which is on a utility pole & 92.50 & 94.05 \\
ml\_photo\_rating\_12\_vs\_45\_v2 & a photo with bad perceptual quality & a photo with good perceptual quality & 83.90 & 80.44 \\
ml\_photo\_rating\_12\_vs\_45\_v3 & a photo with bad perceptual quality & a photo with good perceptual quality & 82.30 & 78.94 \\
ml\_tag\_back\_of\_camera & a photo not showing the back of a camera & a photo showing the back of a camera & 32.30 & 34.06 \\
ml\_tag\_copulation & a photo without copulation & a photo with copulation & 51.50 & 49.10 \\
ml\_tag\_egg & a photo without an egg & a photo with an egg & 76.00 & 77.30 \\
ml\_tag\_foraging\_waterfowl & a photo of a waterfowl not foraging & a photo of a waterfowl foraging & 58.03 & 65.00 \\
ml\_tag\_in\_hand & a photo of a bird which is not in hand & a photo of a bird which is in hand & 96.20 & 97.00 \\
ml\_tag\_nest & a photo without a nest & a photo with a nest & 72.50 & 75.30 \\
ml\_tag\_watermark & a photo without a watermark & a photo with a watermark & 63.10 & 67.90 \\
nabirds\_species\_classification\_amecro\_comrav & a photo of an American Crow & a photo of a Common Raven & 70.25 & 71.06 \\
nabirds\_species\_classification\_bargol\_comgol & a photo of a Barrow's Goldeneye & a photo of a Common Goldeneye & 50.30 & 49.50 \\
nabirds\_species\_classification\_brwhaw\_reshaw & a photo of a Broad-winged Hawk bird & a photo of a Red-shouldered Hawk bird & 91.06 & 91.70 \\
nabirds\_species\_classification\_casvir\_plsvir & a photo of a Cassin's Vireo & a photo of a Plumbeous Vireo & 82.75 & 83.50 \\
nabirds\_species\_classification\_coohaw\_shshaw & a photo of a Cooper's Hawk & a photo of a Sharp-shinned Hawk & 66.00 & 66.90 \\
nabirds\_species\_classification\_easmea\_wesmea & a photo of a eastern meadowlark & a photo of a western meadowlark & 72.40 & 72.25 \\
nabirds\_species\_classification\_semsan\_wessan & a photo of a Semipalmated Sandpiper & a photo of a Western Sandpiper & 62.03 & 63.90 \\
nabirds\_species\_classification\_sursco\_whwsco2 & a photo of a Surf Scoter & a photo of a White-winged Scoter & 49.97 & 51.72 \\
nabirds\_species\_classification\_truswa\_tunswa & a photo of a trumpeter Swan & a photo of a tundra Swan & 70.10 & 70.50 \\
\textbf{Average} &  &  & \textbf{60.25} & \textbf{61.90}
\end{tabular}}
\end{table}

\section{Alternate Training Strategies : Details}

As mentioned in the main paper, our first strategy involves masking texts using ground truth visibility annotations, which are available for the CUB dataset. We manually create a dictionary of mapping of parts to words that might be used to describe the part in the GPT generated texts. For example, \texttt{["leg", "foot", "feet"]} can be used to describe the legs of the bird in the GPT generated texts. Based on this, we mask those sentences for a given image which describe parts for which the gt annotations indicate that the part is not visible. Although this strategy offers improvement (Tab.~\ref{tab:alt} row 2), obtaining these human labelled captions is resource intensive, which is why training using LLM generated texts is useful.

In the second strategy, we assume that CLIP's similarity scores are good indication of visibility. The first way is to pair texts above a threshold of similarity for each image (Tab.~\ref{tab:alt} row 3). We take this threshold as 0.5 probability after taking softmax over texts. The second way is max pooling at instance level (Tab.~\ref{tab:alt} row 4) over image instances and text instances involves generating similarity scores for a small batch of images and texts of the same category to find the pair that has the max similarity and using only that pair to backprop.

For FixMatch (Tab.~\ref{tab:alt} row 5) we generate psuedo soft labels using weaker augmentations and thresholding them with a probability value of 0.5. We finally use the loss we described in \S~\ref{sec:finetune} combined with the cross entropy loss between logits and pseudo labels (with a ratio of 1:0.33). For knowledge distillation (Tab.~\ref{tab:alt} row 6), we use the \S~\ref{sec:finetune} loss combined with the KL divergence (with a ratio 1:4) between the logits of a teacher network scaled by a temperature term (of value 3) and the student logits. For both these methods we initialize every network with pre-trained CLIP. 

\begin{table}[ht]
\centering
\begin{tabular}{lc}
Method & Accuracy \\ \hline
Ours - CLIP\textsuperscript{FT} + A & 54.23 \\
Ground Truth Visibility Masks & \textbf{54.47} \\
Pairing texts with similarity above a treshold & 53.99 \\
Max-pooling at image and text instance level & 54.10 \\
FixMatch & 54.19 \\
Knowledge distillation & 54.38
\end{tabular}
\caption{\textbf{Performance of alternate training strategies trained and tested on CUB dataset.} We use the same train and test class splits as described in the main paper, so as to \textbf{evaluate on unseen classes}. Using GT visibility annotations to mask texts offers improvement. Knowledge distillation offers very little improvement. All others have worse accuracy.}
\label{tab:alt}
\end{table}

\section{Improvement over CLIP-A-Self~\cite{maniparambil2023enhancing}}
CLIP-A-self only tunes a adapter network and lags in performance compared to ours (CLIP\textsuperscript{FT} + A). This again points to the need for fine-tuning CLIP encoders to recognize fine-grained attributes. Another reason why CLIP-A-self suffers is because they query GPT to produce descriptive texts for a fixed set of pre-defined attributes. 
It should also be noted that CLIP-A-self performs worse than CLIP + A in most cases, which shows that even though it does not tune the CLIP encoders, it still overfits to the training classes. We provide results on the all 14 datasets in Tab.~\ref{tab:alldatasets}. For B/16 we compare with CLIP-A-Self in a 16-shot setting for 1:1 train/test split. In Tab.~\ref{tab:cub31} we also compare with 3:1 train/test split for CUB dataset.


\begin{table}[]
\resizebox{\textwidth}{!}{
\begin{tabular}{lccccccccccccccc}
Training Dataset &
  \multirow{2}{*}{CUB} &
  \multirow{2}{*}{FGVC Aircraft} &
  \multirow{2}{*}{Stanford Cars} &
  \multirow{2}{*}{Flowers102} &
  NABirds &
  \multicolumn{2}{c}{iNat} &
  \multirow{2}{*}{EuroSAT} &
  \multirow{2}{*}{Food101} &
  \multirow{2}{*}{ImageNet} &
  \multirow{2}{*}{CalTech101} &
  \multirow{2}{*}{DTD} &
  \multirow{2}{*}{Oxford Pets} &
  \multirow{2}{*}{Sun397} &
  \multirow{2}{*}{UCF101} \\
Testing Dataset  &       &       &       &       & CUB   & CUB   & Flowers &       &       &       &       &       &       &       &       \\ \hline
B/32 CLIP        & 50.54 & 29.27 & 69.72 & 71.78 & 50.54 & 50.54 & 71.78   & 68.89 & 89.94 & 62.61 & 93.87 & 58.45 & 96.64 & 73.03 & 71.12 \\
B/32 CLIP + A     & 50.71 & 30.35 & 69.47 & 75.37 & 50.71 & 50.71 & 75.37   & 72.07 & 90.06 & 65.11 & 94.54 & 59.90 & 96.53 & 75.97 & 75.71 \\
B/32 CLIP\textsuperscript{FT} + A    & \textbf{53.34} & \textbf{36.41} & \textbf{71.63} & \textbf{77.05} & \textbf{55.29} & \textbf{54.58} & \textbf{77.05}   & \textbf{78.56} & \textbf{93.71} & \textbf{65.98} & \textbf{95.75} & \textbf{62.20} & \textbf{96.88} & \textbf{77.75} & \textbf{75.99} \\ \hline
B/16 CLIP        & 51.91 & 36.47 & 74.94 & 77.05 & 51.91 & 51.91 & 77.05   & 64.05 & 92.49 & 67.41 & 93.89 & 60.26 & 97.04 & 75.49 & 77.45 \\
B/16 CLIP + A     & 53.58 & 36.47 & 73.83 & 80.84 & 53.58 & 53.58 & 80.84   & 71.51 & 93.72 & 69.74 & 94.87 & 64.13 & 96.81 & 78.88 & 80.47 \\
B/16 CLIP-A-Self & -     & 33.00 & 72.90 & 75.30 & -     & -     & 75.30   & 70.50 & 91.20 & 68.30 & 95.90 & 62.30 & 97.00 & 76.80 & 76.40 \\
B/16 CLIP\textsuperscript{FT} + A   & \textbf{55.63} & \textbf{40.75} & \textbf{75.78} & \textbf{81.26} & \textbf{56.76} & \textbf{56.77} & \textbf{81.20}   & \textbf{81.82} & \textbf{95.08} & \textbf{71.87} & \textbf{96.03} & \textbf{65.21} & \textbf{97.21} & \textbf{80.32} & \textbf{80.69}
\end{tabular}
}
\caption{\textbf{Results on 14 datasets.} We show performance on ViT B/32 and B/16 with 1:1 train/test class split, obtaining superior accuracy across all settings.}
\label{tab:alldatasets}
\end{table}

\begin{table}[]
\centering
\begin{tabular}{lr}

Method      & \multicolumn{1}{l}{Accuracy} \\ \hline
CLIP        & 71.18                        \\
CLIP + A    & 72.54                        \\
CLIP\textsuperscript{FT} + A  & \textbf{74.80}                        \\
CLIP-A-Self & 71.30                       
\end{tabular}
\caption{\textbf{Performance on CUB with 3:1 train/test split.} Following CLIP-A-Self CUB split we show that we outperform significantly.}
\label{tab:cub31}
\end{table}

\begin{table}[]
\centering
\begin{tabular}{lccc}
 & \multicolumn{1}{l}{CUB} & \multicolumn{1}{l}{FGVC Aircraft} & \multicolumn{1}{l}{Stanford Cars} \\\hline
Multiply Probs & 53.10 & 36.35 & 71.58 \\
Average Probs & 53.34 & 36.41 & 71.63
\end{tabular}
\caption{\textbf{Averaging vs Multiplying Probabilities for aggregating.} Averaging works better than multiplying per-text probabilities for each class for final classification accuracy.}
\end{table}


\section{Correctness of LLM generated texts}
We select 4-6 classes from each of CUB, Stanford Cars and FGVC Aircraft test classes and use all the sentences (about 20 per class) produced by GPT for manual evaluation of correctness. For each sentence participants mark whether an attribute is correct or incorrect. Participants mark some sentences as unsure ($\sim$ 2-4 $\%$). Correctness is determined through various sources across the web such as Wikipedia and All About Birds, by searching the characteristics of the category on Google. Among the sentences marked correct or incorrect over all classes, 96$\%$ (CUB), 90$\%$ (FGVC Aircraft) and 96$\%$ (Stanford Cars) are marked as correct. In FGVC Aircraft where correctness is lower compared to the other two, the incorrect sentences mainly contain measurements in metric units which are usually not easily identifiable visually. For example, the sentence ``An aircraft with a range of approximately 3,600 kilometers." is marked as incorrect as Wikipedia states the number as 3100 kilometers later reduced to 2770 kilometers. We list the sentences for an example class of each dataset with their marked correctness in the following tables.

\begin{table}[hb]
\resizebox{\textwidth}{!}{
\begin{tabular}{llc}
\textbf{Category Name} &
  \textbf{Sentences generated using GPT4} &
  \multicolumn{1}{l}{\textbf{Correctness}} \\ \hline
Downy &
  A bird with a small size, typically measuring between 5.5 to 6.7 inches in length. &
  1 \\
 Woodpecker & A bird with a black and white color pattern.                                               & 1        \\
 & A bird with a white underbelly and lower parts.                                            & 1        \\
 & A bird with black wings that have white spots.                                             & 1        \\
 & A bird with a black tail that has white outer feathers.                                    & 1        \\
 & A bird with a white back.                                                                  & 1        \\
 & A bird with a black head and neck.                                                         & 0        \\
 & A bird with a white stripe above and below the eyes.                                       & 1        \\
 & A bird with a small, pointed beak that is perfect for pecking at wood.                     & 1        \\
 & A bird with a red patch at the back of the head, but only in males.                        & 1        \\
 & A bird with a black bill that is shorter than its head.                                    & 1        \\
 & A bird with a straight, chisel-like bill.                                                  & 1        \\
 & A bird with a large head compared to its body.                                             & 1        \\
 & A bird with a short, stiff tail that provides support against tree trunks.                 & not sure \\
 &
  A bird with a fluttering flight pattern that alternates between flapping and folding its wings. &
  1 \\
 & A bird that is often found in deciduous forests, orchards, parks, and suburban areas.      & 1        \\
 & A bird that is commonly seen alone or in pairs.                                            & 1        \\
 & A bird that is often seen on tree trunks or branches, especially those of deciduous trees. & 1        \\
 & A bird that is frequently seen feeding on insects, seeds, and berries.                     & 1        \\
 & A bird that is native to North America, particularly the United States and Canada.         & 1        \\
 & A bird that is often seen pecking at tree bark in a vertical position.                     & 1        \\
 &
  A bird that is known for its drumming sound on tree trunks, which is a form of communication. &
  1 \\
 & A bird that is often seen in the lower parts of trees or shrubs, unlike other woodpeckers. & 1        \\
 & A bird that is smaller and more delicate than the similar-looking Hairy Woodpecker.        & 1        \\
 &
  A bird that is known for its ability to adapt to human-altered habitats, such as orchards and residential areas. &
  1
\end{tabular}}
\caption{Correctness study of the category ``Downy Woodpecker" of the CalTech-UCSD Birds dataset}
\end{table}

\begin{table}[hb]
\resizebox{\textwidth}{!}{
\begin{tabular}{llc}
\textbf{Category Name} & \textbf{Sentences generated using GPT4}                                                                        & \textbf{Correctness} \\ \hline
Spitfire & An aircraft with an elliptical wing shape.                              & 1 \\
         & An aircraft with a single propeller at the front.                       & 1 \\
         & An aircraft with a long, pointed nose.                                  & 1 \\
         & An aircraft with a large, bubble-shaped cockpit canopy.                 & 1 \\
         & An aircraft with a tail wheel landing gear.                             & 1 \\
         & An aircraft with two exhaust stacks on each side of the engine cowling. & 1 \\
         & An aircraft with a distinctive, rounded vertical stabilizer.            & 1 \\
                       & An aircraft with a Rolls-Royce Merlin or Griffon engine, identifiable by the specific arrangement of exhausts. & 1                    \\
         & An aircraft with a thin, streamlined fuselage.                          & 1 \\
                       & An aircraft with a relatively small horizontal stabilizer compared to the size of the wings.                   & 1                    \\
         & An aircraft with a four-bladed propeller, especially in later models.   & 1 \\
                       & An aircraft with a radiator under each wing in a characteristic rectangular shape.                             & 1                    \\
         & An aircraft with a retractable landing gear.                            & 1 \\
         & An aircraft with a relatively short wingspan compared to its length.    & 0 \\
                       & An aircraft with a distinctive 'kink' in the leading edge of the wing, near the wingtip.                       & 1                    \\
         & An aircraft with a large spinner covering the hub of the propeller.     & 1 \\
                       & An aircraft with a relatively high-set cockpit, giving the pilot a good view.                                  & 1                    \\
         & An aircraft with a narrow track undercarriage.                          & 1 \\
         & An aircraft with a single pilot seat.                                   & 1 \\
         & An aircraft with a distinctive, rounded tailplane.                      & 1
\end{tabular}}
\caption{Correctness study of the category ``Spitfire" of the FGVC Aircraft dataset}
\end{table}

\begin{table}[hb]
\resizebox{\textwidth}{!}{
\begin{tabular}{llc}
\textbf{Category Name} & \textbf{Sentences generated using GPT4}                                           & \textbf{Correctness} \\ \hline
Rolls-Royce & A car with a large, rectangular front grille with vertical slats.                                                     & 1 \\
Ghost Sedan    & A car with a Spirit of Ecstasy hood ornament.                                     & 1                    \\
2012                       & A car with a long, sleek, and luxurious body design.                              & 1                    \\
                       & A car with a high beltline and a low roofline.                                    & 1                    \\
                       & A car with a pair of round LED headlights on each side of the grille.             & 0                    \\
                             & A car with a three-box design, meaning separate compartments for the engine, passenger, and cargo.                    & 1 \\
                       & A car with a large, prominent Rolls-Royce logo on the center of the wheel hubs.   & 1                    \\
                       & A car with a pair of exhaust pipes located at the corners of the rear bumper.     & 1                    \\
                       & A car with a rear-hinged back door, also known as 'suicide doors'.                & 1                    \\
                       & A car with a large, flat hood and a short front overhang.                         & 1                    \\
                       & A car with a long rear overhang, giving it a classic limousine look.              & 1                    \\
                             & A car with a two-tone paint job, often with the hood, roof, and trunk in a different color than the rest of the body. & 1 \\
                       & A car with a large, luxurious, and spacious interior visible through the windows. & 1                    \\
                       & A car with a large, flat trunk lid.                                               & 1                    \\
                             & A car with a distinctive Rolls-Royce clock on the dashboard visible through the windshield.                           & 1 \\
                       & A car with a large, round fuel cap on the right rear side.                        & 0                    \\
                       & A car with a distinctive Rolls-Royce treadplate on the door sill.                 & 1                    \\
                       & A car with a large, rectangular rear window.                                      & 1                    \\
                       & A car with a small, triangular window at the rear of the side windows.            & 1                    \\
                       & A car with a distinctive Rolls-Royce umbrella stored in the rear door.            & 1                   
\end{tabular}}
\caption{Correctness study of the category ``Rolls-Royce Ghost Sedan 2012" of the Stanford Cars dataset}
\end{table}

\end{document}